# PROCESSAMENTO DE LINGUAGEM NATURAL EM PORTUGUÊS E APRENDIZAGEM PROFUNDA PARA O DOMÍNIO DE ÓLEO E GÁS

Diogo da Silva Magalhães Gomes [1,2]
Alexandre Gonçalves Evsukoff [1]

[1] COPPE/Universidade Federal do Rio de Janeiro (UFRJ)
[2] Petrobras, Centro de Pesquisas e Desenvolvimento Leopoldo Américo Miguez de Mello (CENPES)

## Resumo

No decorrer das últimas décadas, instituições em todo o mundo vêm sendo desafiadas a lidar com o imenso volume de informações capturadas em formatos não estruturados. A era da Transformação Digital, caracterizada por importantes avanços tecnológicos e pelo surgimento de métodos disruptivos em técnicas de Inteligência Artificial, oferece oportunidades para promover um melhor aproveitamento dessas informações. Técnicas de Processamento de Linguagem Natural (PLN), principalmente envolvendo recentes abordagens com aprendizagem profunda (*deep learning*), permitem processar de maneira eficiente um grande volume de dados textuais para obtenção de informações relevantes, identificação de padrões, classificação de textos, entre outras aplicações. No contexto do domínio de Óleo e Gás (O&G), entretanto, o vocabulário estritamente técnico da área representa um desafio para esses algoritmos de PLN, cujos termos podem assumir significados muito diferentes do senso comum. A geração de modelos especializados para o domínio demanda a disponibilidade de corpora representativos de O&G em quantidade e qualidade adequadas. Porém, é escasso na literatura científica o acesso público a esse material, especialmente considerando o idioma Português. Este trabalho apresenta uma revisão sobre as principais técnicas disponíveis na área de processamento de linguagem natural e suas aplicações considerando as particularidades do domínio de Óleo e Gás no idioma Português.

## 1 Introdução

Na era da chamada Transformação Digital (HENRIETTE et al., 2017), essencialmente caracterizada pela rica disponibilidade de técnicas de ciência de dados e inteligência artificial, a indústria de Óleo e Gás (O&G) tem sido continuamente desafiada a fazer melhor uso das informações atualmente disponíveis em suas imensas bases de dados (WORLD ECONOMIC FORUM, 2017). Muitas companhias vêm reforçando seus investimentos em tecnologias digitais no intuito de obter um melhor aproveitamento desses dados e, assim, melhorar a eficiência em suas operações e processos decisórios (MATT, HESS e BENLIAN, 2015).

No decorrer das últimas décadas, motivados pela diminuição de custos de *hardware* de armazenamento e processamento, além do surgimento de técnicas e plataformas que tornaram possível processar grandes volumes de dados (Big Data), um novo paradigma se estabeleceu no sentido de armazenar o máximo possível de informações, caracterizando um crescimento exponencial do volume de dados capturados (GARTNER, 2011). Estima-se que uma fração de 80% desses dados sejam armazenados em formatos não estruturados (BLINSTON e BLONDELLE, 2017). O Gartner (2013) define informações não estruturadas como todas aquelas que não atendem a um modelo de dados específico, pré-determinado.

Considerando-se, em especial, documentos no formato texto, tratam-se de relatórios técnicos, artigos científicos, logs de operação, análises laboratoriais, pareceres técnicos, periódicos, entre outros.

Embora essa enorme disponibilidade de dados represente uma oportunidade para as corporações, obter informações relevantes extraídas a partir desses imensos volumes torna-se também um desafio. Muitas instituições não compreendem o potencial de dados que possuem e, portanto, informações essenciais muitas vezes tornam-se negligenciadas por não serem identificadas e processadas adequadamente (FORBES, 2017).

Nesse contexto, recentes avanços nas áreas de Processamento de Linguagem Natural (PLN) e aprendizagem profunda (*deep learning*) vêm contribuindo significativamente para se obter um melhor aproveitamento dessas informações, com o surgimento de novas técnicas para viabilizar o processamento e extração de conhecimento a partir de bases de dados textuais (YOUNG et al., 2018; KHURANA et al., 2018). Há um crescente interesse nessas áreas de conhecimento, o que pode ser confirmado pelo expressivo aumento no número de publicações sobre abordagens com *deep learning*, observado no histórico das principais conferências internacionais especializadas em PLN, conforme relatado por Young et al. (2018) e reproduzido na Figura 1. Essa tendência também está evidenciada pelo Gartner em seu gráfico anual de tecnologias emergentes, onde se encontram mapeadas abordagens associadas a PLN entre as principais tendências, como assistentes virtuais, plataformas de conversação e redes neurais profundas (Figura 2).

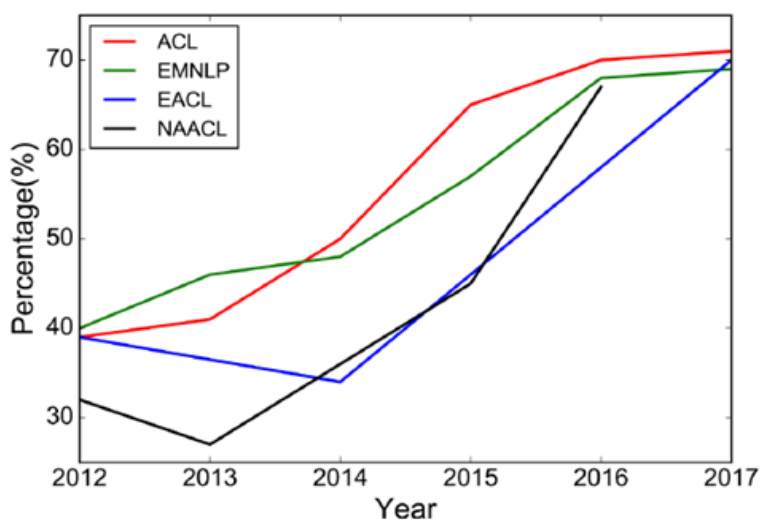

**Figura 1:** *Papers* **com abordagens de** *deep learning* **nas principais conferências de PLN**
**Fonte: Young et al. (2018)**

Algoritmos de PLN, portanto, buscam gerar representações matemáticas significativas para elementos textuais, treinados a partir de conjuntos de dados (*corpora*) representativos do problema a ser processado. Essas representações, por sua vez, tendem a capturar características essenciais de linguagem, como morfologia, sintaxe e, em especial, semântica (MIKOLOV et al., 2013a, 2013b; CAMACHO-COLLADOS e PILEHVAR, 2018). As principais técnicas comumente se fundamentam no princípio de vetorização de palavras, que consiste em atribuir uma representação vetorial densa n-dimensional para cada termo do vocabulário (BENGIO et al., 2013) e são baseadas na hipótese distribucional (SAHLGREN, 2008), caracterizada por assumir que palavras que aparecem em um mesmo contexto tendem a possuir significados semelhantes. Dessa forma, termos relacionados entre si tendem a ser posicionados em uma mesma região de vizinhança no espaço vetorial criado. Essas representações são capazes de otimizar o desempenho de aplicações em PLN devido à sua capacidade de generalização (GOLDBERG, 2016) e são aplicáveis a diversos casos práticos de uso, incluindo-se tradução automática, classificação de texto, sistemas de perguntas e respostas, entre outros (CAMACHO-COLLADOS e PILEHVAR, 2018).

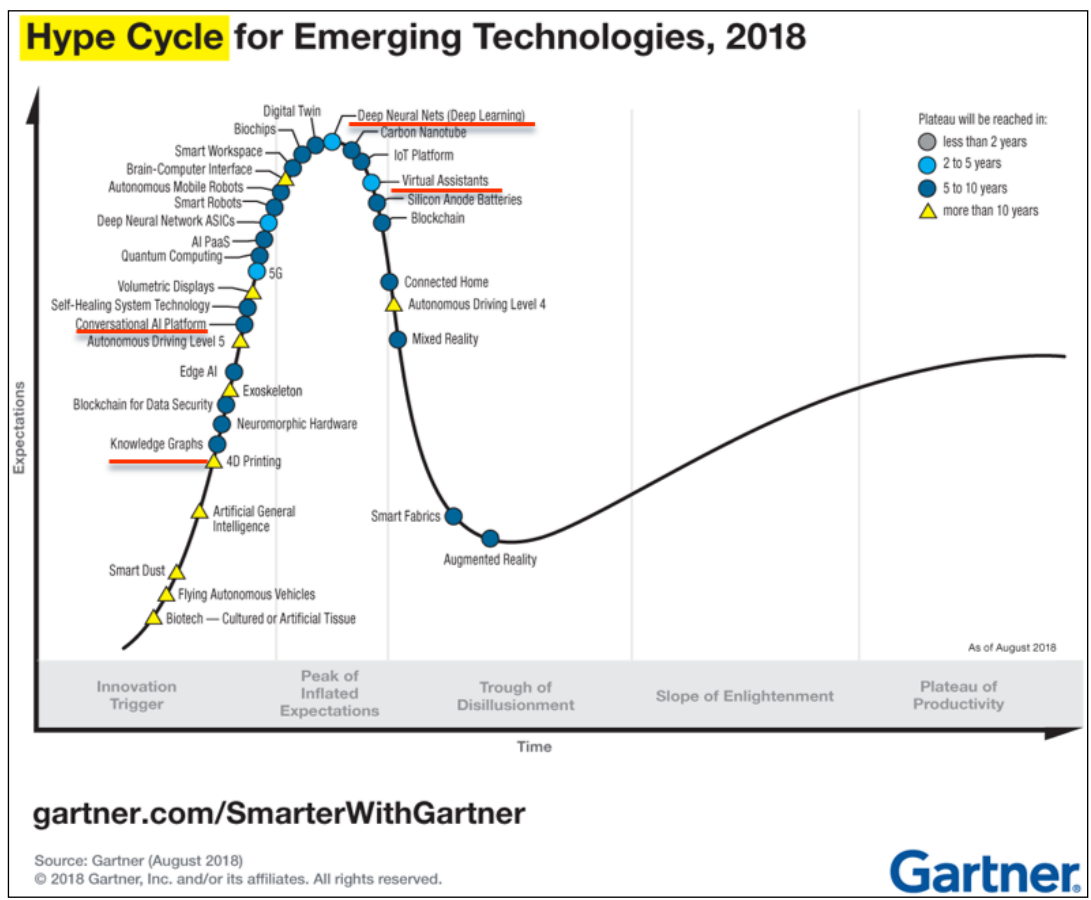

**Figura 2: Principais tecnologias emergentes no ano de 2018**
**Fonte: Gartner (2018)[1].**

      A geração de modelos de PLN especializados demanda a utilização de grandes conjuntos de dados em formato textual (*corpora*), adequadamente preparados para o treinamento dos algoritmos. No entanto, esses dados nem sempre estão disponíveis no contexto de um determinado projeto. Nesse cenário, técnicas conhecidas como transferência de aprendizado (*transfer learning*) são comumente empregadas. Essas técnicas consistem em utilizar modelos pré-treinados a partir de *corpora* de contexto geral, que posteriormente são reutilizados para aplicações em um contexto específico (CER et al., 2018). Alguns dos principais algoritmos recentemente propostos e suas correspondentes implementações disponibilizam modelos pré-treinados em *corpora* de contexto geral (BOJANOWSKI et al., 2017; PENNINGTON et al., 2014). Para o idioma Português, alguns poucos trabalhos se propuseram a conduzir estudos para geração de modelos no idioma (HARTMAN et al., 2017; RODRIGUES et al., 2016). Recentemente, novas técnicas com abordagens usando redes profundas contextuais e direcionais foram propostas, apresentando resultados no estado-da-arte para diferentes aplicações de PLN, tratando também de oferecer modelos pré-treinados para o idioma Português (PETERS et al., 2018; DEVLIN et al., 2018).

## 2   O domínio de Óleo e Gás em Português

      O vocabulário estritamente técnico da área de Óleo e Gás representa um desafio para algoritmos de PLN. Alguns termos podem assumir significados muito distintos em relação ao senso comum, demandando representações mais adequadas para esse domínio específico de forma a manter a correta capacidade de generalização dos modelos (GOMES et al, 2018; NOORALAHZADEH et al., 2018). Adicionalmente, há consistentes evidências que sugerem que a utilização de modelos treinados em *corpora*

---

[1] https://blogs.gartner.com/smarterwithgartner/files/2018/08/PR_490866_5_Trends_in_the_Emerging_Tech_Hype_Cycle_2018_Hype_Cycle.png

específicos do domínio pode melhorar significativamente a qualidade de representação dos termos e, consequentemente, o desempenho dos algoritmos de PLN em que serão utilizados (ALSENTZER, 2019; LAI et al., 2016; DIAZ et al., 2016; NOORALAHZADEH et al., 2018).

A Tabela 1 ilustra alguns exemplos para casos típicos em que termos do vocabulário técnico de O&G assumem um significado completamente distinto de seu senso comum, considerando um contexto geral. Para compor essas definições, selecionamos um conjunto de termos técnicos e utilizamos duas fontes de referência para caracterização de significado: o dicionário Michaelis[2] do idioma Português, para as definições de contexto geral; e o Dicionário do Petróleo[3] para as definições no domínio técnico. Ressalta-se que, para todos os termos, é comum haver mais de uma definição. Portanto, nesses casos, somente a sua definição mais significativa foi considerada. Em tempo, esse aspecto denota características de polissemia, em que um mesmo termo pode assumir diferentes significados, o que representa um desafio adicional aos algoritmos de PLN (CAMACHO-COLLADOS e PILEHVAR, 2018). Complementarmente, cabe destacar ainda os inúmeros termos técnicos que são específicos apenas para a área de O&G, não existindo, portanto, em vocabulários comuns. Esses termos não possuem representação equivalente em modelos de contexto geral, apesar de serem de fundamental importância para o bom desempenho de algoritmos de PLN que precisem considerá-los em suas análises. A Tabela 2 reúne um subconjunto simples de exemplos utilizados com muita frequência em documentos técnicos, selecionados a partir do Siglário[4] publicado pelo Dicionário do Petróleo, para os quais não existe correspondência no Dicionário Michaelis.

**Tabela 1: Definições de termos conforme sua representação no contexto geral e técnico de O&G**

| Termo | Definição – Dicionário Michaelis | Definição: Dicionário do Petróleo |
|---|---|---|
| *Arcósio* | *\* sem definição* | *Rocha sedimentar composta de grãos de tamanho areia e com composição de minerais de feldspatos e quartzo.* |
| *Braço* | *Cada um dos membros superiores do corpo humano.* | *Mola em arco ou alavanca articulada a uma sonda de perfilagem para pressioná-la contra a parede do poço e desse modo manter o patim ou almofada;* |
| *Calado* | *Que não diz nada ou que fica em silêncio; silencioso* | *Distância vertical entre a superfície da água e a parte mais baixa do navio naquele plano, considerando uma embarcação, para um plano transversal de interesse.* |
| *Camisa* | *Peça de vestuário, masculino ou feminino, em geral de tecido leve, com mangas curtas ou compridas, e que se veste ordinariamente sobre a pele e vai desde o pescoço até a altura dos quadris, fechada na frente por uma fileira de botões.* | *Componente da bomba de fundo, utilizado no método de produção por bombeio mecânico, consistindo de um tubo que envolve o pistão e que possui superfície interna polida e perfeitamente ajustada ao pistão para que, durante o ciclo de bombeio, seja minimizado o escorregamento de fluido bombeado.* |
| *Campo* | *Terreno extenso e plano; terreno plano, extenso, com poucas árvores; campina.* | *Área definida por critérios técnico-administrativos que contém uma ou mais acumulações comerciais conhecidas de petróleo em unidades tectônicas, tais como uma bacia sedimentar, ou em geossinclinais, ou seja, em grande bacia geológica que recebeu a sedimentação de grandes espessuras de sedimentos originadas das áreas adjacentes mais elevadas.* |

---

[2] https://michaelis.uol.com.br/moderno-portugues/
[3] http://dicionariodopetroleo.com.br/
[4] http://dicionariodopetroleo.com.br/siglario/

| | | |
|---|---|---|
| *Coque* | *Pancada leve na cabeça com os nós dos dedos; carolo, cascudo.* | *Combustível derivado da aglomeração de carvão e que consiste de matéria mineral e carbono, fundidos juntos.* |
| *DST* | *Sigla de Doença Sexualmente Transmissível* | *Drill Stem Test (Teste de Formação)* |
| *Fadiga* | *Cansaço resultante de trabalho, físico ou intelectual, intenso; fadigação, pregação.* | *Dano progressivo (acumulativo), localizado e permanente que ocorre no material quando a estrutura está submetida a tensões cíclicas.* |
| *Peixe* | *Denominação comum aos animais aquáticos, vertebrados, cujos membros são nadadeiras sustentadas por raios ósseos, com esqueleto ósseo ou cartilaginoso, com pele revestida de escamas, cuja respiração é feita por brânquias.* | *Peça metálica deixada dentro de um poço de petróleo que constitua impedimento ao prosseguimento normal das operações de perfuração.* |
| *Reserva* | *Ação ou efeito de reservar(-se); reservação.* | *Volumes de petróleo e de gás natural estimados como comercialmente recuperáveis pela aplicação de projetos de desenvolvimento em acumulações conhecidas, a partir de uma determinada data, sob condições definidas.* |
| *Reservatório* | *Lugar para armazenar algo; depósito.* | *Configuração geológica dotada de propriedades específicas, armazenadora de petróleo ou gás em subsuperfície.* |
| *Testemunho* | *Narração real e circunstanciada que se faz em juízo; declaração, depoimento.* | *Cilindro de rocha cortado durante a perfuração de poços, que varia normalmente de 2 cm a 25 cm de diâmetro e em vários metros no comprimento.* |
| *Unha* | *Garra recurva e pontiaguda de alguns animais.* | *Acessório utilizado para calçar estruturas.* |

**Tabela 2: Definições de termos técnicos sem equivalência no dicionário Michaelis**

| Termo | Definição: Dicionário do Petróleo |
|---|---|
| *BCS* | *Bombeamento Centrífugo Submerso* |
| *BOP* | *Blowout Preventer (Preventor de Erupção, Obturador de Segurança)* |
| *CLF* | *Conector de Linha de Fluxo (Flowline Hub)* |
| *DHPT* | *Downhole Pressure&Temperature (Temperatura e Pressão no Fundo do Poço)* |
| *FPSO* | *Floating, Production, Storage and Offloading (Unidade Flutuante de Produção, Estocagem e Transferência de Óleo)* |
| *HWCT* | *Horizontal Wet Christmas Tree (Árvore de Natal Molhada Horizontal)* |
| *MMBOE* | *Milhões de Barris de Óleo Equivalente (Million Barrels of Oil Equivalent)* |
| *SMS* | *Segurança, Meio Ambiente e Saúde (Safety, Environment and Health)* |
| *VLCC* | *Very Large Crude Carrier (Navio-tanque de Petróleo)* |

Nesse cenário, a busca por representações adequadas para os termos técnicos, de maneira a promover uma melhor capacidade de generalização dos algoritmos de PLN, demanda a disponibilidade de *corpora* representativo no domínio e idioma. No entanto, apesar da crescente popularização de técnicas de PLN, é escasso na literatura científica o acesso público a modelos e dados de treinamento no domínio específico de Óleo e Gás, especialmente considerando o idioma Português. Uma das primeiras iniciativas nesse sentido foi apresentada por Gomes et al. (2018), disponibilizando para uso público um *corpus* e um conjunto de modelos especializados no domínio. Para o idioma inglês, Nooralahzadeh et al. (2018)

desenvolveram um conjunto de modelos especializados para O&G, assim como uma metodologia para otimização de hiperparâmetros e métricas de avaliação quantitativas dos modelos.

## 3   Revisão de Literatura

A revisão de literatura foi realizada considerando as principais bases disponíveis no catálogo de periódicos da CAPES[5], em especial as bases Science Direct[6] e Scopus[7] (ambas da *Elsevier*), e Web of Science[8] (*Clarivate Analytics*). Para definição dos critérios de pesquisa, foram considerados os principais termos relacionados às áreas de processamento de linguagem natural, aprendizagem profunda e sua contextualização no domínio de Óleo e Gás. Adicionalmente aos termos de busca, a pesquisa foi expandida em função dos autores mais relevantes e com maior número de publicações, identificados a partir das métricas obtidas junto às bases de periódicos. Os artigos foram selecionados em função do impacto de suas contribuições na comunidade científica, por sua aderência ao tema e pela relevância do conteúdo.

Adicionalmente, a fim de contemplar os aspectos relacionados ao domínio de Óleo e Gás, foram também consideradas as bases de publicações especializadas como do Instituto Brasileiro de Petróleo[9] (IBP) e a base OnePetro[10] da *Society of Petroleum Engineers* (SPE).

Além das consultas às bases de periódicos anteriormente mencionadas, cabe considerar que a área de processamento de linguagem natural é contemplada por importantes conferências internacionais, onde estão representados o estado-da-arte e as principais contribuições nesse tema. Portanto, a revisão de literatura contemplou os anais das principais conferências, em especial os organizados pela *Association for Computational Linguistics* (ACL), que disponibiliza a coletânea do material de suas conferências em seu repositório *ACL Anthology*[11]. Nesse sentido, cabe destacar algumas das principais em PLN: *Conference on Computational Natural Language Learning* (CoNLL), *Conference on Empirical Methods in Natural Language Processing* (EMNLP), *Lexical and Computational Semantics and Semantic Evaluation* (SEMEVAL), *North American Chapter of ACL* (NAACL) e *International Conference on Language Resources and Evaluation* (LREC). Em tempo, é relevante notar o expressivo crescimento no número de publicações com abordagens de aprendizagem profunda, conforme relatado por Young et al. (2018) e reproduzido na Figura 1. De uma maneira geral, a área de Inteligência Artificial e seus subdomínios vem apresentando um rápido crescimento em número de participações em conferências (Figura 3), conforme destacado pelo relatório anual especializado da Universidade de Stanford (SHOHAM et al., 2018).

Oportunamente, há ainda uma conferência internacional especializada no estudo do tema voltado para o idioma Português: a *International Conference on the Computational Processing of Portuguese* (PROPOR)[12]. Os anais das edições anteriores estão disponíveis no repositório Springer[13].

A Tabela 3 relaciona a listagem dos principais artigos identificados e analisados na revisão de literatura e que fundamentem os preceitos teóricos sobre o tema, ordenados em função do ano de publicação.

---

[5] http://www.periodicos.capes.gov.br/
[6] https://www.sciencedirect.com/
[7] https://www.scopus.com
[8] https://clarivate.com/products/web-of-science/
[9] https://www.ibp.org.br/publicacoes/
[10] https://www.onepetro.org/
[11] https://aclweb.org/anthology/
[12] http://www.inf.ufrgs.br/propor-2018/
[13] https://link.springer.com/conference/propor

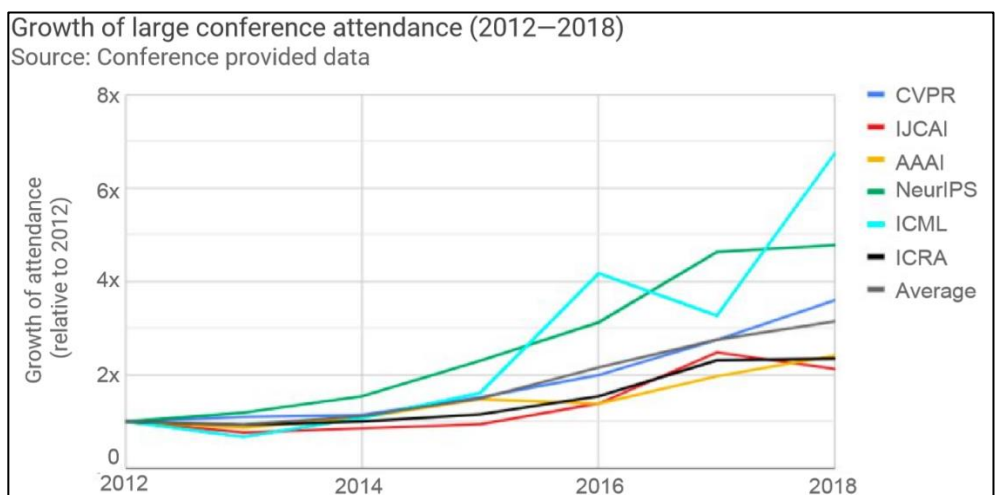

**Figura 3: Crescimento no número de participantes nas principais conferências sobre Inteligência Artificial.**
**Fonte: Shoham et al. (2018).**

**Tabela 3: Principais artigos analisados na revisão de literatura**

| Ano | Autores | Título |
|---|---|---|
| 2019 | Sebastian Ruder | *Neural Transfer Learning for Natural Language Processing* |
| 2019 | Kamran Kowsari, Kiana Jafari Meimandi, Mojtaba Heidarysafa, Sanjana Mendu, Laura Barnes, Donald Brown. | *Text Classification Algorithms: A Survey* |
| 2018 | Diksha Khurana, Aditya Koli, Kiran Khatter, Sukhdev Singh | *Natural Language Processing: State of The Art, Current Trends and Challenges* |
| 2018 | Tom Young, Devamanyu Hazarika, Soujanya Poria, Erik Cambria | *Recent Trends in Deep Learning Based Natural Language Processing* |
| 2018 | Jose Camacho-Collados, Mohammad Taher Pilehvar | *From Word to Sense Embeddings: A Survey on Vector Representations of Meaning* |
| 2018 | Christina Niklaus, Matthias Cetto, André Freitas, Siegfried Handschuh | *A survey on open information extraction* |
| 2018 | Jacob Devlin, Ming-Wei Chang, Kenton Lee, Kristina Toutanova | *Bert: Pre-training of deep bidirectional transformers for language understanding* |
| 2018 | Vikas Yadav, Steven Bethard | *A Survey on Recent Advances in Named Entity Recognition from Deep Learning models* |
| 2018 | Lei Zhang, Shuai Wang, Bing Liu | *Deep Learning for Sentiment Analysis : A Survey* |
| 2018 | Christina Niklaus, Matthias Cetto, André Freitas, Siegfried Handschuh | *A survey on open information extraction* |
| 2018 | Sebastian Ruder, Ivan Vulić, Anders Søgaard | *A Survey of Cross-Lingual Word Embedding Models* |
| 2017 | Nathan S. Hartmann, Erick R. Fonseca, Christopher D. Shulby, Marcos V. Treviso, Jessica S. Rodrigues, Sandra M. Aluísio | *Portuguese Word Embeddings: Evaluating on Word Analogies and Natural Language Tasks* |
| 2017 | Piotr Bojanowski, Edouard Grave, Armand Joulin, Tomas Mikolov | *Enriching Word Vectors with Subword Information* |
| 2017 | Mehdi Allahyari, Seyedamin Pouriyeh, Mehdi Assefi, Saeid Safaei, Elizabeth D. Trippe, Juan B. Gutierrez, Krys Kochut | *Text Summarization Techniques: A Brief Survey* |

| 2016 | Ashwin Ittoo a, Le Minh Nguyen, Antal van den Bosch | *Text analytics in industry: Challenges, desiderata and trends* |
|---|---|---|
| 2016 | Yoav Goldberg | *A primer on neural network models for natural language processing* |
| 2016 | Siwei Lai, Kang Liu, Liheng Xu, Jun Zhao | *How to Generate a Good Word Embedding* |
| 2016 | Hannah Bast, Björn Buchhold, Elmar Haussmann | *Semantic Search on Text and Knowledge Bases* |
| 2015 | Yann LeCun, Y. Bengio, Geoffrey Hinton | *Deep Learning* |
| 2015 | Rafal Jozefowicz, Wojciech Zaremba, Ilya Sutskever | *An Empirical Exploration of Recurrent Network Architectures* |
| 2014 | Jeffrey Pennington, Richard Socher, Christopher D. Manning | *Glove: Global vectors for word representation* |
| 2013 | Tomas Mikolov, Kai Chen, Greg Corrado, Jeffrey Dean | *Efficient Estimation of Word Representations in Vector Space* |
| 2013 | Tomas Mikolov, Ilya Sutskever, Kai Chen, Greg Corrado, Jeffrey Dean | *Distributed Representations of Words and Phrases and their Compositionality* |
| 2008 | Ronan Collobert, Jason Weston | *A Unified Architecture for Natural Language Processing: Deep Neural Networks with Multitask Learning* |

## 3.1  Pesquisas Relacionadas

A fim de analisar tendências e estimular novas descobertas em assuntos correlacionados ao objeto de estudo principal, foram analisadas variações de critérios de consultas utilizando as ferramentas Google Trends[14], Google Acadêmico[15] e Microsoft Academic[16], além dos recursos de *analytics* oferecidos pelas bases Scopus e Web of Science. Essas pesquisas contribuíram para fornecer uma visão mais ampla do espectro de aplicações das técnicas sendo analisadas, inclusive em outros domínios, como nas áreas de direito, linguística, biomédica, ciência da computação e engenharias (Figura 4 e Figura 5).

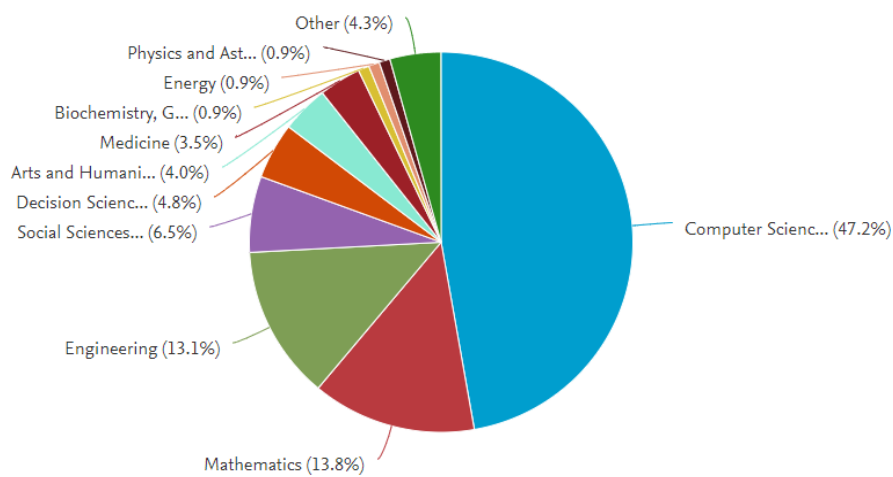

**Figura 4: Publicações referentes aos termos *"nlp"* e *"deep learning"* em diferentes áreas**
**Fonte: Scopus.**

---

[14] https://trends.google.com/trends/
[15] https://scholar.google.com.br/
[16] https://academic.microsoft.com/

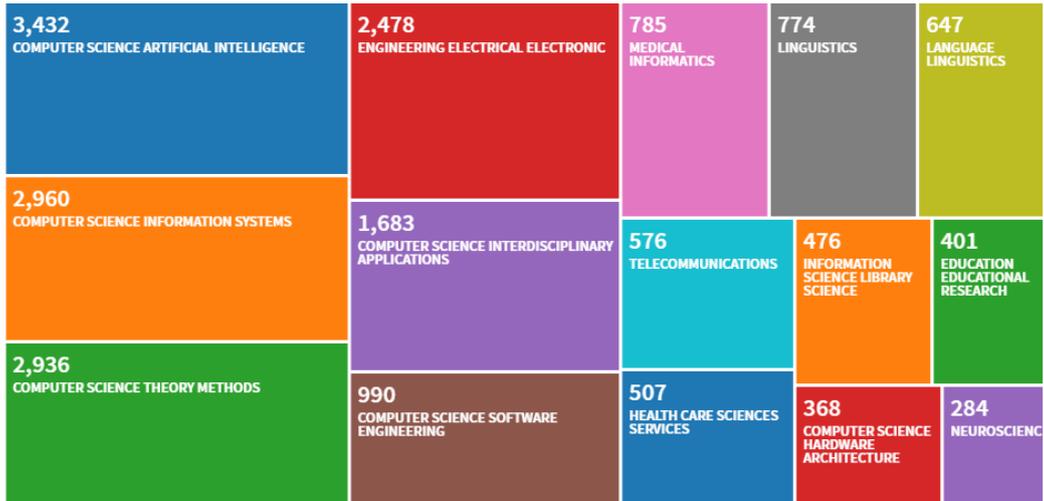

**Figura 5: Publicações referentes ao tópico *"natural language processing"* em diferentes áreas**
**Fonte: Web of Science.**

Primeiramente, com o objetivo de reforçar a relevância do tema, foi analisada a evolução do número de publicações nessa área, utilizando como critério de busca os termos *"deep learning"* e *"natural language processing"*. Observa-se o grande interesse nessa área de conhecimento, evidenciado pelo salto no número de publicações sobre esse tópico, tanto na base Scopus quanto na base Web of Science, conforme ilustrado na Figura 6, Figura 7 e Figura 8.

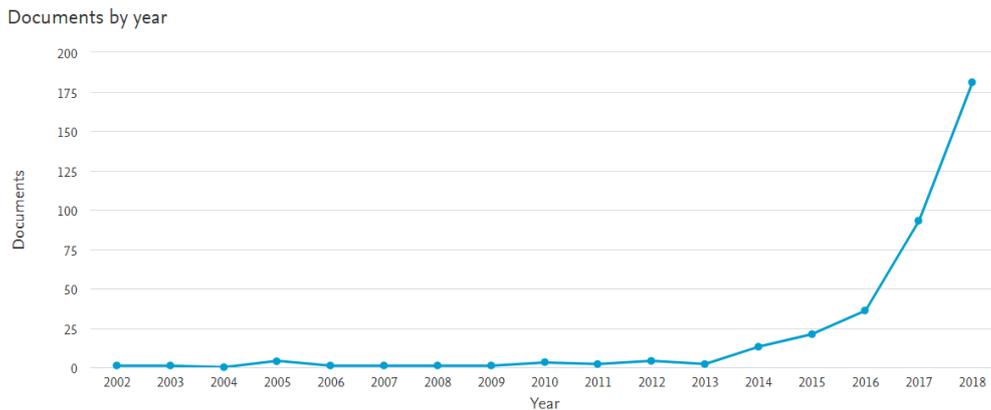

**Figura 6: Publicações referentes aos termos combinados *"nlp"* e *"deep learning"***
**Fonte: Scopus.**

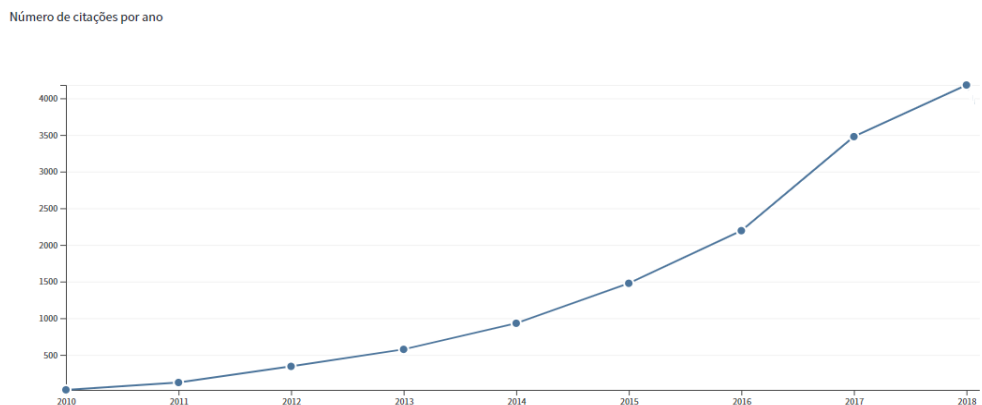

**Figura 7: Número de citações sobre o tópico *"natural language processing"* por ano**
**Fonte: Web of Science.**

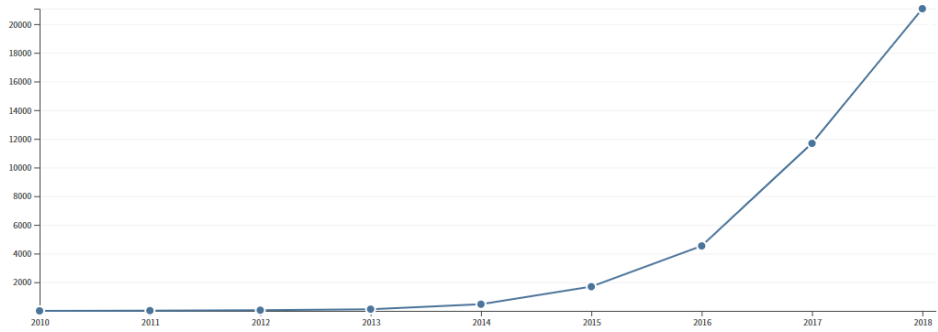

**Figura 8: Número de citações sobre o tópico *"deep learning"* por ano**
**Fonte: Web of Science.**

Em seguida, foi analisado o cenário atual com as principais publicações nessa área de conhecimento, identificando os *journals*, autores, instituições e conferências mais influentes nesse tema. Para isso, foi utilizado o Microsoft Academic para gerar relatórios analíticos sobre o tópico *"natural language processing"*, conforme ilustrado na Figura 9. Esses resultados auxiliaram no refinamento dos critérios de pesquisa bibliográfica, fornecendo parâmetros mais relevantes para buscar as contribuições de estado-da-arte nessas tecnologias.

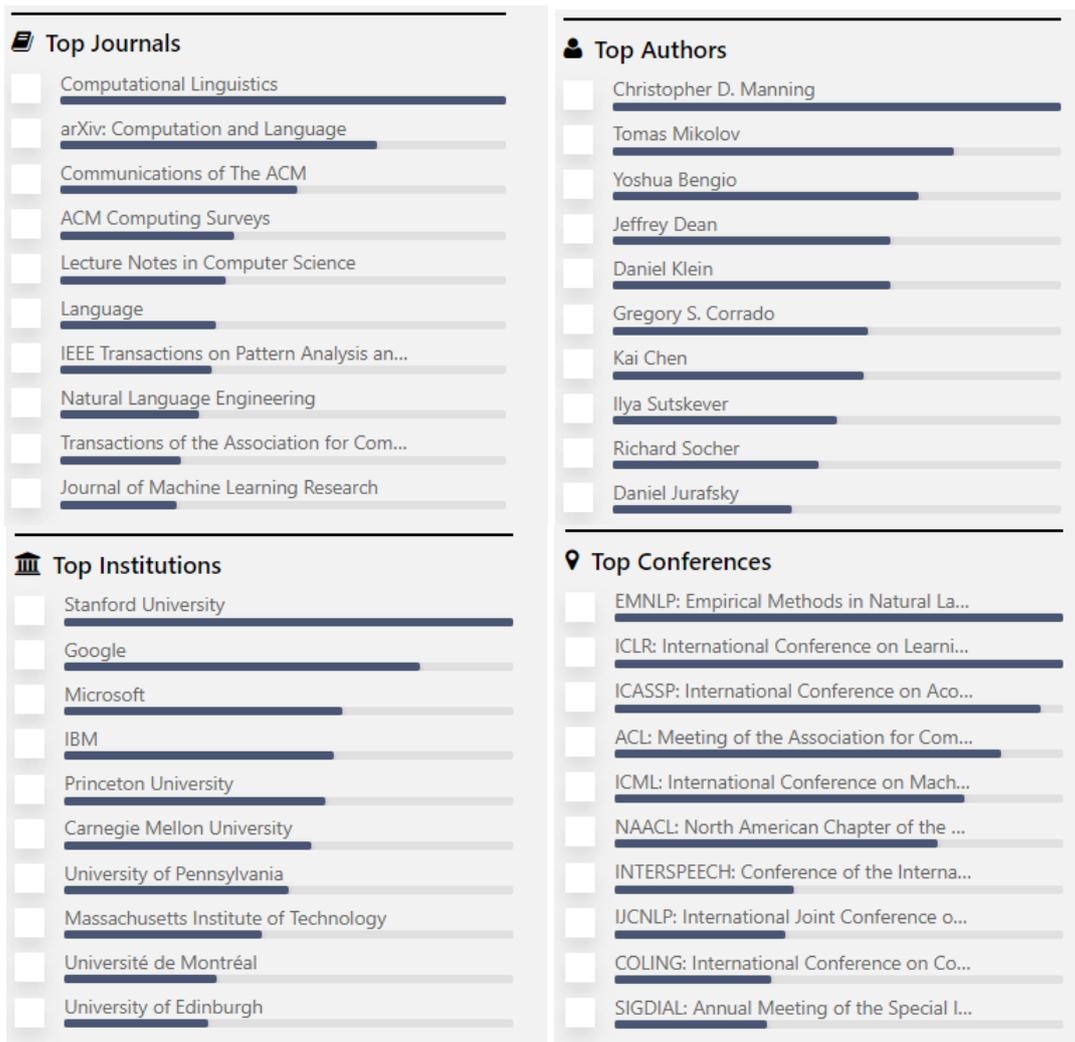

**Figura 9: *Journals*, autores, instituições e conferências mais influentes sobre *"NLP"***
**Fonte: Microsoft Academic.**

Em publicações especializadas no domínio de Óleo e Gás, nota-se um expressivo número de artigos com abordagens associadas aos tópicos *"natural language processing"* e *"deep learning"*, reforçando a relevância dessas técnicas aplicadas a esse domínio. A Figura 10 ilustra o resultados da pesquisa bibliográfica a partir da base OnePetro, considerando esses tópicos.

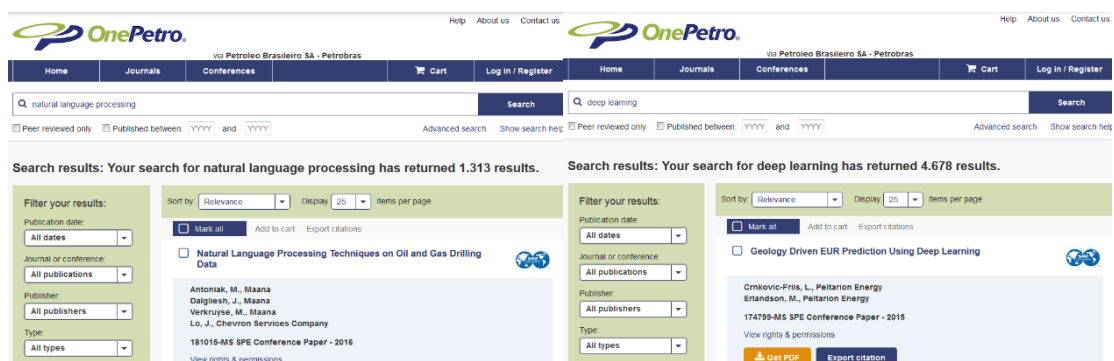

**Figura 10: Publicações sobre os tópicos *"natural language processing"* e *"deep learning"* em *journals* especializados no domínio de O&G**
**Fonte: OnePetro.**

## 4   Processamento de Linguagem Natural

Nas últimas décadas, a indústria e a comunidade científica presenciaram uma expansão sem precedentes na quantidade de informações capturadas e armazenadas em formatos não estruturados, destacando-se especialmente aquelas do tipo textuais. Observou-se a evolução de um cenário inicialmente dominado por informações estruturadas, armazenadas em sistemas gerenciadores de bases de dados (SGBDs) e *data warehouses*, para a situação atual em que formatos não-estruturados são dominantes (ITTOO et al., 2016). Estima-se que uma fração de aproximadamente 80% dos dados são armazenados em formatos não estruturados (BLINSTON e BLONDELLE, 2017).

Dispersas nesses imensos volumes, há importantes informações que, se corretamente identificadas e processadas, podem fornecer relevantes insumos para inúmeros processos decisórios, além de dar suporte a uma ampla variedade de atividades de pesquisa e industriais (ITTOO et al., 2016).

Diante desse cenário, há uma grande demanda para o desenvolvimento de técnicas e mecanismos que permitam fazer um uso mais adequado do imenso potencial econômico, acadêmico e estratégico das informações armazenadas nesses repositórios. Nesse contexto, diversas técnicas na área de PLN se apresentam com um promissor potencial para viabilizar essas tarefas.

A disciplina de processamento de linguagem natural é um ramo da área de conhecimento em inteligência artificial que tem por objetivo fazer com que o computador possa entender e processar palavras e sentenças escritas em linguagem humana. Seus estudos iniciais remontam da década de 1950, resultantes da interseção dos primeiros esforços nos campos de inteligência artificial e linguística (KHURANA et al., 2017). Originalmente, havia distinções conceituais entre PLN e a área de recuperação da informação (*information retrieval*) (MANNING et al., 2008). No entanto, a evolução dessas disciplinas convergiu ao longo do tempo, de maneira que a área de PLN demanda do pesquisador conhecimentos em uma ampla diversidade de especializações (NADKARNI et al., 2011), como linguística computacional, estatística e aprendizagem automática (*machine learning*).

Oportunamente, no contexto de linguística, cabe definir alguns termos comumente utilizados no contexto de trabalhos em PLN: fonologia é o ramo destinado ao estudo de sons na linguagem; morfologia dedica-se à estrutura interna de formação das palavras, a partir da composição de suas unidades básicas (morfemas); sintaxe se refere à relação estrutural das palavras na formação das sentenças; enquanto semântica se refere à compreensão e sentido do texto (KHURANA et al., 2018).

As abordagens iniciais em PLN tradicionalmente se fundamentavam em análise segmentada (*parsing*) e em bases de regras manualmente elaboradas. No entanto, a natureza essencialmente ambígua da forma de expressão da linguagem e a necessidade de identificar, em uma última instância, a semântica das palavras, ultrapassam a capacidade humana em elaborar bases de regras com uma cobertura adequada (ITTOO et al., 2016). Portanto, é desejável que algoritmos de PLN sejam capazes de resolver automaticamente estrutura sintática, desambiguações de palavras e compreender o escopo semântico de uma sentença (MANNING e SCHÜTZE, 1999).

Essa necessidade deu origem à ascensão de novas abordagens de aprendizagem automática, focadas na identificação de padrões a partir de dados de exemplo e fortemente baseadas em métodos estatísticos. Nesse cenário, a área de PLN evoluiu com a aplicação de técnicas como, por exemplo, *Support Vector Machines* (SVM), *Hidden Markov Models* (HMM) e *Conditional Random Fields* (CRF) (ITTOO et al., 2016). Dessa forma, inicialmente essa área foi dominada por abordagens de aprendizagem automática baseadas em métodos lineares, treinados a partir de vetores esparsos com muitas dimensões. No entanto, essas arquiteturas se deparavam com os altos custos computacionais para o processamento desses vetores, e comumente esbarravam com a chamada "maldição da dimensionalidade" (*curse of dimensionality*) (BENGIO et al., 2003).

Recentemente, houve uma ampla proposição de técnicas de forma a adotar modelos neurais não-lineares baseados em vetores densos como dados de entrada, em substituição aos modelos lineares baseados em vetores esparsos (GOLDBERG, 2016). Uma das principais contribuições nessa área se trata do uso de representações distribuídas de palavras (*distributed word representations*), que se caracterizam por representar palavras e conceitos através de vetores densos e com um número de dimensões predefinido (MANNING, 2015). Estas técnicas capacitaram os algoritmos de aprendizagem profunda a capturar representações de similaridade entre palavras, permitindo assim promover uma generalização mais adequada durante suas etapas de treinamento. Manning (2015) destaca a importância que essas técnicas representam para algoritmos de inteligência artificial, que precisam ser capazes de inferir o significado de um conceito ou expressão a partir de suas menores partes constituintes (as palavras). Nesse sentido, Young et al. (2018) descrevem que grande parte do sucesso obtido pelos modelos neurais e, mais recentemente, pelas abordagens com aprendizagem profunda (*deep learning*), é atribuído às contribuições fornecidas pelo advento das representações distribuídas de palavras (também conhecidas como *word embeddings*), que deram suporte à maior capacidade de generalização dos algoritmos e, por conseguinte, ao exponencial avanço da área de PLN.

No entanto, é a partir do ano de 2015 que se torna evidente na área de PLN a forte tendência de utilização de métodos baseados em aprendizagem profunda (LECUN, BENGIO e HINTON, 2015; GOODFELLOW, BENGIO e COURVILLE, 2016), notadamente observável por sua massiva presença nas principais conferências especializadas (conforme anteriormente ilustrado na Figura 1). Nesse ano, conforme relatado por Manning (2015), eminentes pesquisadores na área de *deep learning* publicaram sobre suas visões de futuro, reforçando a importância dessas abordagens cada vez mais presentes para superar os desafios de PLN. Em especial, cabe destacar as relevantes opiniões publicadas por Yann Lecun[17], Geoffrey Hinton[18] e Yoshua Bengio.

## 4.1 Representações Distribuídas

Antes de analisar mais detalhadamente arquiteturas neurais de PLN, é importante compreender a estrutura fundamental utilizada como unidade de entrada para esses algoritmos. Em linguagem natural, podemos considerar a palavra como sendo essencialmente uma unidade básica de significado. Portanto, é necessário estabelecer representações matemáticas adequadas para os dados de entrada em formato texto, de forma que seja possível processá-los em algoritmos de aprendizagem automática.

Inicialmente, abordagens tradicionais utilizavam a técnica conhecida como *one-hot encoding*, que consiste em utilizar representações categóricas em vetores esparsos, de forma que cada propriedade (*feature*) é representada por uma dimensão desse vetor (GOLDBERG, 2016). Ou seja, cada termo é representado como um vetor $\mathbb{R}^{|V|\times 1}$ preenchido por zeros (onde |V| é o tamanho do vocabulário), contendo somente uma posição com o valor 1 no índice correspondente à palavra no vocabulário. No entanto, em PLN os vocabulários de referência normalmente são muito grandes (ocasionalmente alcançando milhões de palavras [19]), o suficiente para ocasionar o problema da explosão da dimensionalidade, tornando computacionalmente ineficiente o processamento de vetores muito grandes.

---

[17] http://www.wired.com/2014/12/fb/
[18] https://www.reddit.com/r/MachineLearning/comments/2lmo0l/ama_geoffrey_hinton

Adicionalmente, para que os algoritmos de PLN consigam uma capacidade de generalização adequada, é desejável conhecer certas noções de similaridade entre as palavras. Como a técnica *one-hot* representa cada palavra como uma entidade totalmente independente, não há nenhuma relação entre os vetores de cada termo. Ou seja, quaisquer pares de palavras terão seus vetores sempre ortogonais e equidistantes entre si (Equação 1), não guardando, portanto, quaisquer informações sobre seus significados.

$$(v^{óleo})^T v^{gás} = 0 = (v^{cachorro})^T v^{cão}$$

**Equação 1: Vetores *one-hot* ortogonais**

Nesse sentido, um dos principais saltos obtidos na transição dos modelos lineares para os modelos neurais diz respeito à substituição das representações esparsas, em que cada *feature* é representada por uma dimensão do vetor (vetores *one-hot*), por representações densas de vetores contínuos com dimensões predefinidas (*embeddings*) (GOLDBERG, 2015). Ou seja, cada propriedade é embutida em um espaço vetorial n-dimensional e mapeada como um vetor nesse espaço.

Além da evidente melhoria computacional de processamento, uma das principais vantagens das representações distribuídas está capacidade de capturar diversas relações de similaridade entre as palavras nas dimensões embutidas dos vetores, permitindo uma melhor generalização dos modelos. Essa propriedade de generalização dos modelos vetoriais, ao serem integradas em arquiteturas baseadas em redes neurais, foi comprovadamente decisiva para o atingimento de resultados no estado-da-arte em diversas aplicações de PLN (CAMACHO-COLLADOS e PILLEVAR, 2018; YOUNG et al., 2018; HARTMAN et al., 2017; KHURANA et al., 2017; GOLDBERG, 2016; LAI et al., 2016; SCHNABEL et al., 2015).

Representações distribuídas de palavras (*distributed word representations*) são normalmente atribuídas através de contexto, induzidas a partir de dados textuais de exemplo através de métodos de treinamento não-supervisionados. A década de 1990 é marcada por importantes contribuições nessa área (ELMAN, 1991), que estabeleceram as fundações nesse campo de pesquisa e que evoluiu com o advento das técnicas de *Latent Semantic Analisys* (LSA) e modelos de linguagem (BENGIO et al., 2003). Outras técnicas seguiram como adaptações das propostas anteriores, que levaram à criação de disciplinas de modelagem de tópicos (*topic modelling*), baseadas em técnicas como *Latend Dirichlet Allocation* (BLEI, NG e JORDAN, 2003).

Algoritmos de vetorização de palavras (*word embeddings*) processam grandes conjuntos de dados textuais (*corpus*), obtendo representações em vetores densos de valores reais, capazes de capturar características essenciais de linguagem como sintaxe e semântica (MIKOLOV et al., 2013a, 2013b; HARTMAN et al., 2017), com grande capacidade de generalização (GOLDBERG, 2016). Essas técnicas são baseadas na hipótese distribucional (SAHLGREN, 2008), que consiste em considerar que palavras com significados semelhantes tendem a aparecer em um mesmo contexto. Dessa forma, termos similares tendem a ter seus vetores posicionados em uma mesma região de vizinhança no espaço vetorial criado. Portanto, essas relações de similaridade podem ser inferidas a partir do cálculo da distância entre seus vetores. A distância cosseno atua como uma métrica de similaridade entre dois termos considerando o ângulo entre seus vetores (Equação 2).

---

[19] https://code.google.com/archive/p/word2vec/

$$\cos(\theta) = \frac{\mathbf{A} \cdot \mathbf{B}}{\|\mathbf{A}\|\|\mathbf{B}\|} = \frac{\sum\limits_{i=1}^{n} A_i B_i}{\sqrt{\sum\limits_{i=1}^{n} A_i^2}\sqrt{\sum\limits_{i=1}^{n} B_i^2}},$$

**Equação 2: Similaridade Cosseno**

Collobert e Weston (2008) apresentaram o primeiro estudo demonstrando a utilidade prática de vetores de palavras pré-treinados, estabelecendo a técnica de *word embeddings* como uma referência para aplicações de PLN. O alto índice de utilização e de citações a essas técnicas na literatura científica recente mostra sua efetividade e sua importância em qualquer modelo de *deep learning* desempenhando tarefas em PLN (YOUNG et al., 2018).

No entanto, uma das principais contribuições para a popularização dessa abordagem é o advento do Word2vec (MIKOLOV et al., 2013a, 2013b), que apresentou técnicas baseadas em redes neurais computacionalmente eficientes para treinamentos dos vetores em grande escala, além de *datasets* e métricas de avaliação que evidenciaram sua grande capacidade de generalização ao capturar essas propriedades de similaridade sintática e semântica. O Word2vec propõe duas diferentes arquiteturas de treinamento: *continuous bag-of-words* (CBOW) e *skip-gram* (Figura 11). O modelo CBOW objetiva predizer a palavra central a partir do conjunto de palavras vizinhas dentro de uma janela de contexto de tamanho *k*. Ou seja, o objetivo do treinamento é maximizar a probabilidade de observar a palavra atual, considerando como entrada as palavras de contexto. O *skip-gram*, por sua vez, realiza o oposto, objetivando predizer a distribuição (probabilidade) das palavras de contexto a partir de uma palavra central de referência (YOUNG et al., 2018). A função objetivo do *skip-gram* consiste em minimizar o somatório dos erros de predição entre todas as palavras na camada de saída (MIKOLOV et al., 2013a).

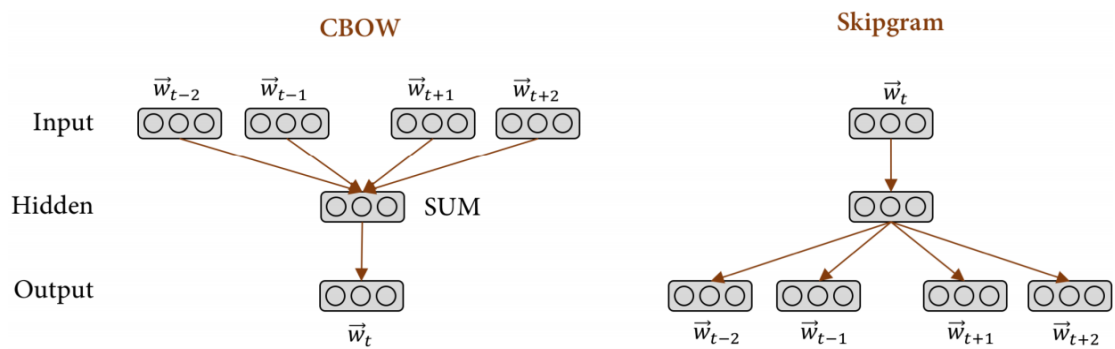

**Figura 11: Ilustração dos diferentes modelos propostos pelo Word2vec: CBOW e skip-gram
Fonte: Camacho-Collados e Pillevar (2018).**

Mikolov et al. (2013a) evidenciaram, inclusive, que a realização de operações algébricas entre vetores sugere que as relações semânticas entre os termos são mantidas (Figura 12). Em seu clássico exemplo, a seguinte operação vetorial resulta em uma posição no espaço vetorial que atende à igualdade: $v^{Paris} - v^{France} + v^{Italy} = v^{Rome}$, e apresentou ainda outras relações de analogias semânticas válidas (Figura 13).

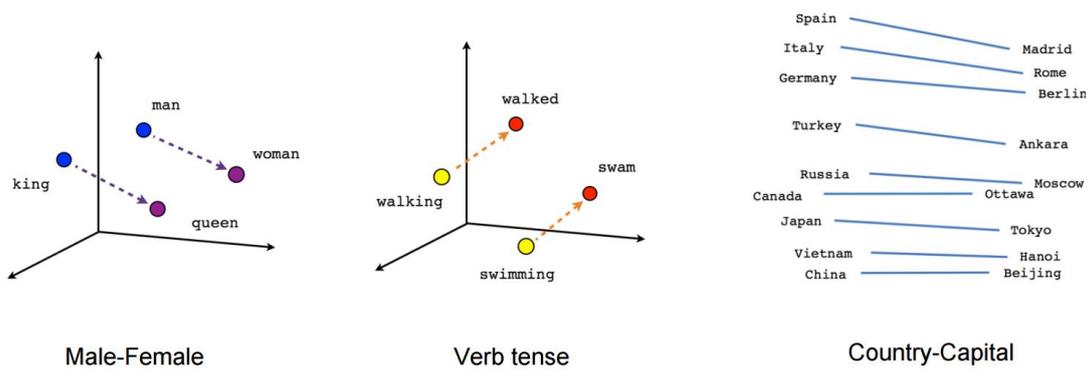

**Figura 12: Operações vetoriais nos *embeddings* com manutenção de suas relações semânticas
Fonte: Tensorflow[20].**

| Relationship | Example 1 | Example 2 | Example 3 |
|---|---|---|---|
| France - Paris | Italy: Rome | Japan: Tokyo | Florida: Tallahassee |
| big - bigger | small: larger | cold: colder | quick: quicker |
| Miami - Florida | Baltimore: Maryland | Dallas: Texas | Kona: Hawaii |
| Einstein - scientist | Messi: midfielder | Mozart: violinist | Picasso: painter |
| Sarkozy - France | Berlusconi: Italy | Merkel: Germany | Koizumi: Japan |
| copper - Cu | zinc: Zn | gold: Au | uranium: plutonium |
| Berlusconi - Silvio | Sarkozy: Nicolas | Putin: Medvedev | Obama: Barack |
| Microsoft - Windows | Google: Android | IBM: Linux | Apple: iPhone |
| Microsoft - Ballmer | Google: Yahoo | IBM: McNealy | Apple: Jobs |
| Japan - sushi | Germany: bratwurst | France: tapas | USA: pizza |

**Figura 13: Relações de analogias semânticas a partir de operações vetoriais nos *embeddings*
Fonte: Mikolov et al. (2013a).**

Na sequência, outras técnicas de vetorização de palavras foram propostas. *Global Vectors* (GloVe) (PENNINGTON et al., 2014) é um método de vetorização em que o treinamento é realizado a partir de estatísticas globais de co-ocorrência entre as palavras. Segundo Pennington et al. (2014), o GloVe apresenta resultados consistentemente melhores do que o Word2vec em tarefas de analogias semânticas.

Recentemente, Bojanowski et al. (2017) apresentaram o FastText, uma variação da arquitetura Word2vec em que os *embeddings* são associados a sequências contíguas de caracteres (*n-grams*), e cada palavra é representada pelo somatório das representações de seus *n-grams*. Essa característica lhe atribui a particular capacidade de permitir representações inclusive para termos não observados no vocabulário durante o treinamento, além de capturar características morfológicas na formação dos termos.

Levy et al. (2015) apresentam uma detalhada pesquisa comparando diferentes abordagens de vetorização de palavras, inclusive Word2vec e Glove. O artigo apresenta orientações para otimização dos hiperparâmetros, sugerindo que estes possuem grande impacto na qualidade dos vetores, enquanto não foram observadas grandes variações de desempenho entre os diferentes algoritmos, contrastando com estudos anteriores.

## 4.2 Representações Contextuais

As representações distribuídas de palavras incorrem em algumas limitações ao assumir um vetor único e universal para cada termo. Isso implica em desconsiderar características essenciais de linguagem como a polissemia, não sendo capazes de representar diferentes significados que uma mesma palavra pode assumir. Essa característica é conhecida como *meaning conflation deficiency*, em que os diferentes significados da palavra são projetados em um único ponto do espaço vetorial. Na tentativa de representar unicamente os diferentes significados de um termo, isso pode acarretar o efeito de atrair para seu entorno outras palavras não-semanticamente relacionadas, que estariam por sua vez associadas a regiões representadas pelos diferentes significados do termo (NEELAKANTAN et al., 2014).

---

[20] https://www.tensorflow.org/tutorials/word2vec

Representações contextuais de palavras assumem uma nova perspectiva na atribuição de vetores para cada termo de referência, permitindo variar conforme o contexto em que o termo ocorre. Ou seja, o vetor de uma palavra pode mudar dinamicamente entre o processamento de uma sentença e outra, dependendo do contexto.

Desde o advento das técnicas de *word embeddings*, a abordagem tradicionalmente utilizada para implementar algoritmos de PLN consiste em inicializar a primeira camada com modelos de vetores pré-treinados (como Word2vec, Glove, entre outros), enquanto o restante da rede é treinado para uma finalidade específica. Nesse sentido, diferentes abordagens propõem uma nova estratégia, que consiste em pré-treinar toda a arquitetura de rede profunda e transferir todas as suas camadas, especializando apenas as camadas de topo ao serem reaplicados em outra tarefa (RUDER, 2019).

Peters et al. (2018) propuseram o *Embedding from Language Model* (ELMo), que produz diferentes *embeddings* para cada contexto em que uma palavra é referenciada, permitindo, portanto, diferentes representações para uma mesma palavra. Adicionalmente, utiliza representações internas baseadas em caracteres, permitindo considerar características morfológicas dos termos e torna possível a representação de palavras fora do vocabulário de treinamento. Há modelos públicos pré-treinados disponíveis no repositório do projeto[21], inclusive para o idioma Português.

Em sequência, outras propostas foram apresentadas, como *Universal Language Model Fine-tuning* (ULMFit) (HORWARD e RUDER, 2018), endereçando especialmente problemas de classificação, sendo possível sua reutilização em outras tarefas utilizando *transfer learning*.

Esses dois métodos baseiam-se no conceito de modelos de linguagem (*language models*) em sua etapa de pré-treinamento, que consistem em computar a probabilidade de ocorrência de um conjunto de palavras em uma sequência particular (MANNING e SOCHER, 2017). Uma das limitações dessa abordagem é não considerar simultaneamente o contexto tanto à direita como à esquerda da palavra-alvo. Nesse sentido, uma nova proposta recentemente apresentada por Devlin et al. (2018) é o *Bidirectional Encoder Representations from Transformers* (BERT), que estabeleceu novos parâmetros em estado-da-arte para diferentes *benchmarks* de PLN. Esse modelo baseia-se na arquitetura *Transformer* (VASWANI et al., 2017), e considera o contexto das palavras à direita e à esquerda de forma verdadeiramente bidirecional em todas as camadas. Como resultado, novas aplicações podem fazer uso de modelos BERT pré-treinados[22] otimizando apenas uma camada adicional para especializá-los a outra tarefa específica.

# 5   Principais Aplicações de PLN

No contexto das diversas técnicas relacionadas a problemas de PLN, cabe destacar algumas das aplicações mais comuns nessa área de conhecimento, conforme abaixo listadas e descritas nos parágrafos a seguir:

- Tradução automática;
- Reconhecimento de entidades Nomeadas (REN)
- Classificação automática de textos;
- Análise de sentimentos;
- Sistemas de perguntas e respostas;
- Extração da informação;
- Sumarização automática;
- Busca semântica.

---

[21] https://allennlp.org/elmo
[22] https://github.com/google-research/bert#pre-trained-models

A área de tradução automática concentra uma das primeiras iniciativas na linha de pesquisa em PLN, tendo mantido sua relevância até os dias atuais, especialmente considerando a intensa necessidade de comunicação promovida pela globalização das relações e pelas plataformas de redes sociais. Uma das mais importantes contribuições recentemente publicadas nessa área foi o uso de mecanismos de atenção com a arquitetura *Transformer* (VASWANI et al., 2017). Ruder, Vulic e Søgaard (2018) apresentam uma detalhada pesquisa sobre o estado-da-arte em *word embeddings* multi-idiomas.

O Reconhecimento de Entidades Nomeadas (REN, *Named Entity Recognition*) objetiva identificar entidades (nomes próprios) contidas em um texto. Algumas das principais categorias tradicionalmente consideradas para as entidades são: pessoa, local, organização e tempo (LAMPLE et al., 2016), enquanto domínios específicos podem definir seus próprios conjuntos de entidades. Yadav e Bethard (2018) apresentam uma visão geral sobre REN baseados em modelos de aprendizagem automática.

A classificação automática de texto corresponde a uma das mais fundamentais tarefas em PLN, e objetiva atribuir uma ou mais categorias pré-estabelecidas para um determinado texto. Está diretamente relacionada com outras aplicações, como detecção de *spam*, análise de sentimentos e identificação de idioma. Recentemente, métodos de aprendizagem automática vêm aumentando sua popularidade nesse tipo de aplicação, notadamente em função de sua capacidade de abstrair modelos complexos e as relações não-lineares inerentes a esses dados. As abordagens predominantes adotam arquiteturas baseadas em redes recorrentes ou redes convolucionais, embora outras alternativas sejam também experimentadas. Kowsari et al. (2019) apresentam uma ampla pesquisa sobre o histórico e o estado-da-arte nessa área de aplicação.

A análise de sentimentos visa aplicar algoritmos computacionais para classificar a polaridade de opiniões relacionadas a uma afirmação. Ocasionalmente, processa grandes volumes de informação, em que os resultados são normalmente classificados como positivos, negativos ou neutros, podendo haver, entretanto, outras categorias. Essas técnicas têm sido amplamente empregadas em contextos comerciais e industriais. Zhang et al. (2018) apresentam um detalhado estudo sobre abordagens de aprendizagem profunda para problemas de análise de sentimento.

Sistemas de perguntas e respostas (*question answering*) objetivam analisar uma pergunta formulada em linguagem humana e determinar sua resposta. Comumente atuam em um domínio restrito, de maneira que os sistemas de PLN podem explorar o conteúdo de *corpora* do domínio específico na construção de suas bases de conhecimento. O *dataset* academicamente mais relevante nessa área é o *Stanford Question Answering Dataset* (SQuAD)[23] (RAJPURKAR et al., 2016). Há uma estreita relação desta área com a construção de assistentes virtuais para atendimento automatizado (*dialogue systems*), com um imenso potencial de interesse comercial e industrial, conforme anteriormente evidenciado no gráfico de tecnologias emergentes do Gartner (**Erro! Fonte de referência não encontrada.**). Uma visão sobre diferentes técnicas e abordagens utilizadas nessas áreas é apresentado por Young et al. (2018).

Extração da informação (*information extraction*) busca extrair conteúdo relevante e estruturado a partir de conteúdos textuais, comumente restritos a um determinado domínio. Faz uso de outras técnicas como reconhecimento de entidades nomeadas (NER), resolução de correferências e extração de relações, entre outros. Niklaus et al. (2018) apresentam uma visão sobre o estado-da-arte nessa área.

Algoritmos de sumarização automática objetivam reconhecer em um texto os seus trechos mais significativos e o que pode ser descartado, no intuito de compor um resumo legível do seu conteúdo mais relevante. Um estudo sobre essa área de aplicação é apresentado por Allahyari et al. (2017)

Busca semântica (*semantic search*) consiste no uso de mecanismos mais inteligentes para melhor compreender a intenção do usuário ao formular seus critérios de pesquisa por determinada informação. Objetiva considerar o significado da *query*, indo além de uma simples consulta por palavras-chave indexadas, buscando agregar as diferentes variações de representação incluídas em um mesmo espectro semântico de um determinado conceito. Bast et al. (2016) apresentam uma extensa abordagem a respeito de busca semântica em textos e bases de conhecimento.

---

[23] https://rajpurkar.github.io/SQuAD-explorer/

# 6   Aprendizagem profunda para processamento de linguagem natural

No contexto dos recentes avanços na área de processamento de linguagem natural, notadamente impulsionados pelo sucesso obtido pelo uso de diferentes abordagens de aprendizagem profunda, cabe detalhar alguns dos principais conceitos e algoritmos disponíveis. Em linhas gerais, há três principais arquiteturas neurais que se tornaram predominantes em aplicações de PLN: redes recorrentes, redes convolucionais e redes recursivas. No entanto, novas abordagens têm apresentado potencial promissor, conforme detalhado ao longo desta subseção.

Redes neurais recorrentes (RNN) (ELMAN, 1990) objetivam executar o processamento dos dados de entrada em sequência, de maneira que um determinado passo recebe como entrada o resultado do processamento do passo anterior (Figura 14). RNNs, uma vez desdobradas, podem ser entendidas como se fossem redes *feed-forward* muito profundas, em que todas as suas camadas compartilham os mesmos pesos (LECUN, BENGIO e HINTON, 2015). Esse compartilhamento dos pesos ao longo de vários passos no processamento sequencial permite o aprendizado de padrões que ocorrem em posições distintas da cadeia de entrada, conferindo propriedades que lhe atribuem analogias ao conceito de "memória". RNNs representam uma escolha natural para modelar problemas em PLN, considerando sua capacidade de capturar as características inerentemente sequenciais presentes na linguagem, processadas em cadeias de caracteres, palavras ou sentenças (Young et al., 2018).

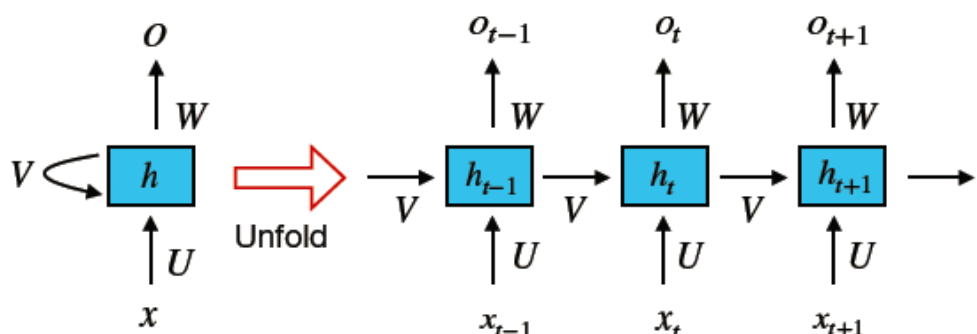

**Figura 14: RNN desdobrada em passos iterativos**
**Fonte: Lecun, Bengio, Hinton (2015).**

As primeiras abordagens neurais para modelos de linguagens apresentados por Bengio et al. (2001) baseadas em redes *feed-foward* foram substituídas por arquiteturas baseadas em redes recorrentes (MIKOLOV et al., 2010). No entanto, RNNs são difíceis de serem treinadas para identificar dependências em posições muito distantes, em função do problema da explosão ou desaparecimento do gradiente (BENGIO et al., 2013). Portanto, a arquitetura original das RNN comumente empregadas em PLN logo evoluiu para adoção de redes *long short-term memory* (LSTM) (HOCHREITER e SCHMIDHUBER, 1997), em função de sua melhor robustez ao lidar com o problema do gradiente no decorrer das camadas profundas. Redes LSTM introduzem explicitamente o conceito de células de memória, responsáveis por preservar o gradiente por uma longa cadeia de sequências (LECUN, BENGIO e HINTON, 2015), além de incluir uma porta adicional para "esquecimento" (GERS, SCHMIDHUBER e CUMMINS, 1999). A Figura 15 ilustra um diagrama de uma rede LSTM.

Cho et al. (2014) propuseram uma arquitetura alternativa para simplificar o complexo funcionamento das células LSTM, apresentando a unidade *Gated Recurrent Unit* (GRU), em que uma única porta é responsável por controlar simultaneamente o fator de esquecimento e a decisão de atualizar sua célula interna. Jozefowicz et al. (2015) apresentam um detalhado estudo comparativo sobre diferentes arquiteturas de redes recorrentes, incluindo redes LSTM e GRU, explorando suas potencialidades e fraquezas em diferentes cenários.

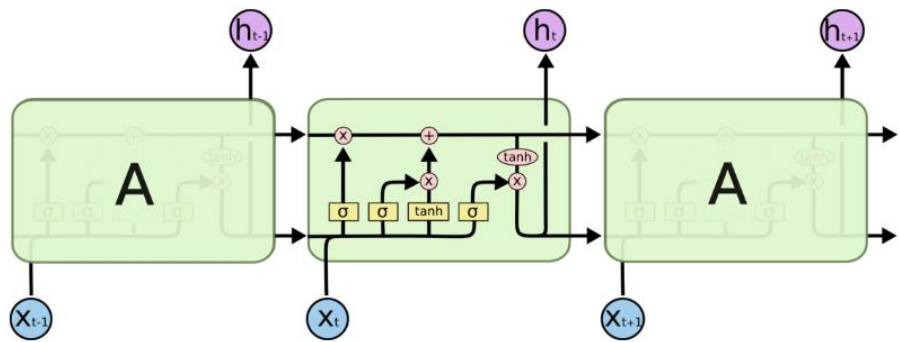

**Figura 15: Rede LSTM**
**Fonte: Olah (2015).**

Redes neurais convolucionais (CNN) se tornaram populares em função de seu sucesso na área de visão computacional (KRIZHEVSKY, SUTSKEVER e HINTON, 2012), e rapidamente foram adaptadas para aplicações em processamento de linguagem natural (KIM, 2014). Não obstante, Collobert e Weston (2008) estão entre os primeiros pesquisadores a registrar a aplicação de CNN em tarefas de PLN.

Em arquiteturas CNN, os dados de entrada (caracteres, palavras ou sentenças) são representados como vetores (*embeddings*), que podem ser iniciados de forma aleatória ou transferidos a partir de pré-treinamento – entretanto, utilizar vetores pré-treinados em grandes *corpora* pode contribuir para melhoria do desempenho (KIM, 2014). Filtros convolucionais (*kernels*) são então aplicados, deslocando-se conforme uma janela de tamanho *n*. Essa operação é repetida em diversas camadas, com variações nos tamanhos dos filtros, intercalando com operações de *max-pooling* para reduzir a dimensão e aumentar o nível de abstração dessas representações. Nesse contexto, em PLN os filtros convolucionais operam iterando ao longo da dimensão temporal representada pelas cadeias de entrada. A Figura 16 ilustra tipicamente a utilização de uma CNN em PLN.

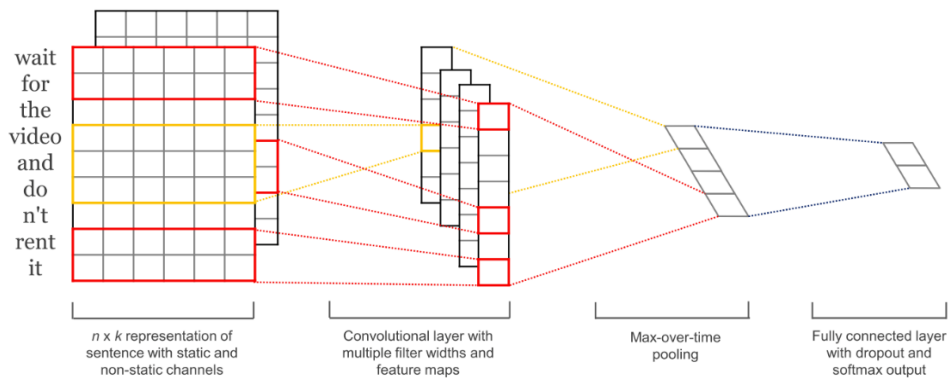

**Figura 16: Rede convolucional para processamento de texto**
**Fonte: Kim (2014).**

Uma das vantagens das redes CNN sobre as RNN é sua maior capacidade de paralelização, uma vez que cada passo da convolução depende apenas de seu contexto local, enquanto as RNNs dependem de toda a cadeia de estados anteriores. Nesse contexto, Yin et al. (2017) apresentam um estudo comparativo sobre o desempenho de CNN e RNN em diferentes aplicações de PLN, e concluem não haver uma opção única definitivamente melhor, dependendo, portanto, das condições globais e semânticas envolvidas em cada tarefa específica. Um detalhado levantamento das principais arquiteturas e seus resultados nas correspondentes aplicações de PLN é apresentado por Young et al. (2018).

Arquiteturas RNN consideram o processamento do texto subdividindo-o como uma sequência de dados de entrada. Entretanto, também é possível compreender a linguagem segundo uma perspectiva hierárquica, onde as palavras são combinadas de maneira recursiva para compor expressões que, por sua vez, podem ser novamente combinadas em uma ordem mais alta para compor novos conceitos. Essa ideia inspirou a interpretação de sentenças como árvores hierárquicas ao invés de sequências (SOCHER et al., 2013).

Redes neurais recursivas (RecNN) constroem a representação de suas sentenças segundo uma perspectiva de baixo para cima (*bottom-up*), em contraste com redes recorrentes que consideram como uma sequência da direita para a esquerda ou esquerda para direita (YOUNG et al, 2018). A representação em uma estrutura de árvore permite que os modelos recursivos possam fazer melhor uso da interpretação sintática da estrutura das sentenças, sendo particularmente útil em situações de negação, por exemplo, em que uma palavra produz efeito direto em todo um segmento da sentença, conforme ilustrado na Figura 17.

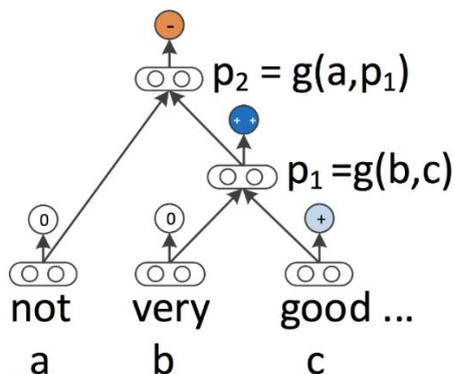

**Figura 17: Estrutura da rede recursiva aplicada a classificação de sentimentos**
**Fonte: Socher et al. (2013).**

Recentemente, outras propostas surgiram como evolução das principais arquiteturas anteriormente abordadas. Sutskever et al. (2014) apresentam uma abordagem sequência-a-sequência baseada em arquitetura neural codificador-decodificador (*encoder-decoder*). Uma rede recorrente é usada para codificar a sequência de entrada como uma representação $y_n$, e esse vetor é usado como entrada auxiliar para uma segunda RNN responsável pela decodificação (GOLDBERG, 2015). Essa arquitetura se mostrou imediatamente adequada para problemas de tradução automática (SUTSKEVER et al., 2014), sendo aplicada em implementações de tradução automática (*neural machine translation*) pelo Google (WU et al., 2016) alcançando resultados de estado-da-arte.

Uma das principais dificuldades em modelos de sequencia-a-sequencia está no fato de compactarem toda a entrada em um único vetor de tamanho fixo (*encoder*), incluindo informações que podem não ser totalmente relevantes no contexto da tarefa atual ou nos casos em que a entrada é muito longa, que depois será expandido novamente (*decoder*). Nesse sentido, o mecanismo de atenção (BAHDANAU, CHO e BENGIO, 2015) busca aliviar esse problema permitindo que o *decoder* tenha acesso a informações da sequência de entrada. Ou seja, durante a decodificação, além do último estado, o *decoder* também tem acesso a um vetor de contexto calculado em função da sequência de entrada (YOUNG et al., 2018), tornando possível inferir trechos relevantes que merecem mais ou menos atenção no processamento. Uma variação dessa abordagem é conhecida como auto-atenção (*self-attention*) que pode ser utilizada para analisar palavras vizinhas na sentença para obter mais informações contextuais. Vaswani et al. (2017) propôs a arquitetura Transformer, estruturada por múltiplas camadas empilhadas de *self-attention*, que representa atualmente o estado-da-arte em tradução automática.

## 7  Conclusão

Este trabalho apresentou uma revisão sobre as principais técnicas de processamento de linguagem natural e suas aplicações considerando as particularidades do domínio específico de Óleo e Gás no idioma Português, com especial ênfase nas abordagens baseadas em aprendizagem profunda. O domínio de O&G carece de conjuntos de dados textuais públicos academicamente estabelecidos que possam fundamentar o desenvolvimento de modelos especializados. Nesse sentido, cabe destacar as iniciativas descritas no trabalho de Gomes et al. (2018), para a disponibilização de corpora e modelos vetoriais especializados.

# Referências

ALLAHYARI, M.; POURIYEH, S. A.; ASSEFI, M.; SAFAEI, S., TRIPPE, E., GUTIERREZ, J., KOCHUT, K. **Text Summarization Techniques: A Brief Survey**. 2017. International Journal of Advanced Computer Science and Applications (IJACSA). 8. 397-405. 10.14569/IJACSA.2017.081052.

ALSENTZER, Emily et al, Publicly Available Clinical BERT Embeddings, 2019. Proceedings of the 2nd Clinical Natural Language Processing Workshop, Association for Computational Linguistics.

BAHDANAU, D., CHO, K., BENGIO, Y. (2015). Neural Machine Translation by Jointly Learning to Align and Translate. **CoRR**, abs/1409.0473.

BARONI, M., DINU, G., KRUSZEWSKI, G. Don't count, predict! a systematic comparison of context-counting vs. context-predicting semantic vectors. In **ACL**, pages 238–247.

BAST, H., BUCHHOLD, B., HAUSSMANN, E. (2016). **Semantic Search on Text and Knowledge Bases**. Foundations and Trends® in Information Retrieval. 10. 119-271. doi:10.1561/1500000032.

BENGIO, Y., DUCHARME, R., VINCENT, P., JANVIN, C. (2003). A Neural Probabilistic Language Model. **The Journal of Machine Learning Research**, 3, 1137-1155.

BENGIO, Y., SIMARD, P., FRASCONI, P. Learning long-term dependencies with gradient descent is difficult. **IEEE Trans. Neural Networks 5**, 157–166 (1994).

BLEI, D., NG, A., JORDAN, M., Latent dirichlet allocation. **Journal of machine Learning research**, vol. 3, no.Jan, pp. 993–1022, 2003

BLINSTON, K., BLONDELLE, H. Machine learning systems open up access to large volumes of valuable information lying dormant in unstructured documents. The Leading Edge, 2017.

BOJANOWSKI, P., GRAVE, E., JOOULIN, A. MIKOLOV, T. Enriching Word Vectors with Subword Information. **Transactions of the Association for Computational Linguistics**. 2017.

CAMACHO-COLLADOS, J., PILEHVAR, M. From Word to Sense Embeddings: A Survey on Vector Representations of Meaning. **Journal of Artificial Intelligence Research** 63, 2018.

CER, D., YANG, Y., KONG, S.-Y., HUA, N., KIMTIACO, N., JOHN, R. S., CONSTANT, N., GUAJARDO-CESPEDES, M., YUAN, S., TAR, C., SUNG, Y.-H., STROPE, B., KURZWEIL, R. Universal sentence encoder. ArXiv, 1803.11175, 2018.

CHO, K.; MERRIENBOER, B. van; GULCEHRE, C.; BAHDANAU, D.; BOUGARES, F.;SCHWENK, H.; BENGIO, Y. Learning phrase representations using rnn encoder–decoder forstatistical machine translation. In:EMNLP. [S.l.]: Association for Computational Linguistics, 2014. p. 1724–1734.

COLLOBERT, R.; WESTON, J. **A Unified Architecture for Natural Language Processing: Deep Neural Networks with Multitask Learning**. Proceedings of the 25th International Conference on Machine Learning. ICML '08.New York, NY, USA: ACM, 2008.

DEVLIN, J., CHANG, M.-W., LEE, K. TOUTANOVA, K. Bert: Pre-training of deep bidirectional transformers for language understanding. arXiv preprint arXiv:1810.04805, 2018.

DIAZ, F., MITRA, B., CRASWELL, N. Query Expansion with Locally-Trained Word Embeddings. Proceedings of the 54th Annual Meeting of the Association for Computational Linguistics, p. 367–77. doi:10.18653/v1/P16-1035, Berlin, Germany, 2016

ELMAN, J. L. Finding structure in time. **Cognitive Science**, v. 14, n. 2, p. 179–211, 1 abr. 1990.

ELMAN, J. L. Distributed Representations, Simple Recurrent Networks, And Grammatical Structure. **Machine Learning - Connectionist approaches to language learning**., v. 7, n. 2–3, p. 195–225, set. 1991.

FORBES, 2017. The Big (Unstructured) Data Problem. https://www.forbes.com/sites/forbestechcouncil/2017/06/05/the-big-unstructured-data-problem/#67629e6e493a

GARTNER, 2018. 5 Trends Emerge in the Gartner Hype Cycle for Emerging Technologies. Disponível em: <https://www.gartner.com/smarterwithgartner/5-trends-emerge-in-gartner-hype-cycle-for-emerging-technologies-2018/>. Acesso em: 24 jul. 2019.

GARTNER, 2011. *Information Management Goes 'Extreme': The Biggest Challenges for 21st Century CIOs*. [online]. Disponível em: http://togetherwepass.co.za/http://togetherwepass.co.za/wp-content/uploads/filebase/certified_office_manager/management/study_notes_/Extreme-Information-Management%20Extreme.pdf (Acessado em março/2019)

GARTNER, 2013. https://blogs.gartner.com/darin-stewart/2013/05/01/big-content-the-unstructured-side-of-big-data/

GERS, F., SCHMIDHUBER, J., CUMMINS, F. Learning to forget: Continual prediction with lstm, 9th International Conference on Artificial Neural Networks, pp. 850–855, 1999.

GOLDBERG, Y. (2016). A primer on neural network models for natural language processing. **Journal of Artificial Intelligence Research**, 57, 345-420.

GOMES, D., CORDEIRO, F., EVSUKOFF, A. 2018. Word Embeddings em Português para o Domínio Específico de Óleo e Gás. **In Proceedings of Rio Oil & Gas Expo and Conference 2018**.

GOODFELLOW, I., BENGIO, Y., COURVILLE, A. **Deep Learning**. MIT Press. Disponivel em: <http://www.deeplearningbook.org>, 2016.

HARTMAN, N. S., FONSECA, E., SHULBY, C., TREVISO, M., RODRIGUES, J. S., ALUISIO, S. Portuguese Word Embeddings: Evaluating on Word Analogies and Natural Language Tasks. Proceedings of Symposium in Information and Human Language Technology. Uberlandia, MG, Sociedade Brasileira de Computacão, 2017.

HENRIETTE, E.; FEKI, M.; BOUGHZALA, I.. The Shape of Digital Transformation: A Systematic Literature Review. **MCIS 2015 Proceedings**, 2015. Disponível em: <https://aisel.aisnet.org/mcis2015/10>.

HOCHREITER, S.; SCHMIDHUBER, J. Long Short-Term Memory. **Neural Computation**, v. 9, n. 8, p. 1735–1780, nov. 1997.

HOWARD, J.; RUDER, S. Universal language model fine-tuning for text classification. arXiv preprint arXiv:1801.06146, 2018.

ITTOO, A.; NGUYEN, L. M.; VAN DEN BOSCH, A. Text analytics in industry: Challenges, desiderata and trends. **Computers in Industry**, Natural Language Processing and Text Analytics in Industry. v. 78, p. 96–107, 1 maio 2016.

JOZEFOWICZ, R., ZAREMBA, W., SUTSKEVER, I. An Empirical Exploration of Recurrent Network Architectures. In: **Proceedings of the 32Nd International Conference on International Conference on Machine Learning - Volume 37**. [s.l.]: JMLR.org, 2015, p. 2342–2350. (ICML'15).

KHURANA, Diksha; KOLI, Aditya; KHATTER, Kiran., SINGH, S. (2017). Natural Language Processing: State of The Art, Current Trends and Challenges. *CoRR abs/1708.05148*.

KOWSARI, K., MEIMANDI, K., HEIDARYSAFA, M., MENDU, S., BARNES, L., BROWN, D. (2019) "Text Classification Algorithms: A Survey". **Information 10**, nº 4 (abril de 2019): 150. https://doi.org/10.3390/info10040150.

KRIZHEVSKY, A.; SUTSKEVER, I.; HINTON, G. E. **ImageNet Classification with Deep Convolutional Neural Networks**. Proceedings of the 25th International Conference on Neural Information Processing Systems - Volume 1. NIPS'12.USA, 2012.

KIM, Y. Convolutional Neural Networks for Sentence Classification. . In: PROCEEDINGS OF THE 2014 CONFERENCE ON EMPIRICAL METHODS IN NATURAL LANGUAGE PROCESSING (EMNLP). 2014

LAI, S., Liu, K., XU, L., ZHAO, J. How to Generate a Good Word Embedding. IEEE Intelligent Systems, vol. 31, no. 6, pp. 5-14. doi: 10.1109/MIS.2016.45. Nov.-Dec. 2016.

LAMPLE, G.; BALLESTEROS, M.; SUBRAMANIAN, S., KAWAKAMI, K., DYER, C**. Neural Architectures for Named Entity Recognition**. In: Proceedings of the 2016 Conference of the North American Chapter of the Association for Computational Linguistics: Human Language Technologies. San Diego, California: Association for Computational Linguistics, 2016, p. 260–270. Disponível em: <http://aclweb.org/anthology/N16-1030>. Acesso em: 9 abr. 2019.

LE, Q., MIKOLOV, T. Distributed Representations of Sentences and Documents. In: **International Conference on Machine Learning**. [s.l.: s.n.], 2014.

LECUN, Y., BENGIO, Y., HINTON, G. Deep Learning. **Nature**, maio 2015.

LEVY, O.; GOLDBERG, Y.; DAGAN, I. Improving Distributional Similarity with Lessons Learned from Word Embeddings. **Transactions of the Association for Computational Linguistics**, v. 3, p. 211–225, 2015.

MANNING, C., SOCHER, R. Lecture Notes on Natural Language Processing with Deep Learning. Stanford University, 2017. Disponível em: <https://web.stanford.edu/class/archive/cs/cs224n/cs224n.1184/syllabus.html>

MANNING, C. Computational Linguistics and Deep Learning. **Computational Linguistics**, v. 41, n. 4, p. 701–707, 2015.

MANNING, C. D.; RAGHAVAN, P.; SCHÜTZE, H. **Introduction to information retrieval**. New York: Cambridge University Press, 2008.

MANNING, Christopher D.; SCHÜTZE, Hinrich. **Foundations of statistical natural language processing**. Cambridge, Mass: MIT Press, 1999.

MATT, Christian; HESS, Thomas; BENLIAN, Alexander. Digital Transformation Strategies. **Business & Information Systems Engineering**, v. 57, n. 5, p. 339–343, 2015.

MIKOLOV, T., CHEN, K., CORRADO, G., DEAN, J. Efficient estimation of word representations in vector space. ICLR Workshop, 2013a.

MIKOLOV, T., SUTSKEVER, I., CHEN, K., CORRADO, G., DEAN, J. Distributed representations of words and phrases and their compositionality. Proceedings of the 26th International Conference on Neural Information Processing Systems-Volume 2(NIPS'13). 2013b.

MIKOLOV, T., KARAFIÁT, M., BURGET, L., CERNOCKÝ, J., KHUDANPUR, S. (2010). Recurrent neural network based language model. Proceedings of the 11th Annual Conference of the International Speech Communication Association, INTERSPEECH 2010. 2. 1045-1048.

MUNEEB, T. H., SAHU, S. K., ANAND, A. Evaluating distributed word representations for capturing semantics of biomedical concepts. Workshop on Biomedical Natural Language Processing (BioNLP). 2015.

NADKARNI, P. M.; OHNO-MACHADO, L.; CHAPMAN, W. W. Natural language processing: an introduction. **Journal of the American Medical Informatics Association: JAMIA**, v. 18, n. 5, p. 544–551, out. 2011.

NEELAKANTAN, A. SHANKAR, J.; PASSOS, A. McCallum, A**.. Efficient Non-parametric Estimation of Multiple Embeddings per Word in Vector Space**. Proceedings of the 2014 Conference on Empirical Methods in Natural Language Processing (EMNLP). Anais... In: PROCEEDINGS OF THE 2014

CONFERENCE ON EMPIRICAL METHODS IN NATURAL LANGUAGE PROCESSING (EMNLP). Doha, Qatar: Association for Computational Linguistics, 2014

NIKLAUS, C., CETTO, M., FREITAS, A., HANDSCHUH, S. **A survey on open information extraction**. CoRR, abs/1806.05599, 2018.

NOORALAHZADEH, F., ØVRELID, L., LØNNING, J. 2018. Evaluation of Domain-specific Word Embeddings using Knowledge Resources. LREC (2018).

OLAH, C. Understanding LSTM Networks. 2015 Disponível em: <https://colah.github.io/posts/2015-08-Understanding-LSTMs/>. Acesso em: 15 abr. 2019.

PENNINGTON, J, SOCHER, R., MANNING, C. D. Glove: Global vectors for word representation, in Empirical Methods in Natural Language Processing (EMNLP), pp. 1532–1543.http://www.aclweb.org/anthology/D14-1162, 2014.

PETERS, M., NEUMANN, M., IYYER, M., GARDNER, M., CLARK, C., LEE, K., ZETTLEMOYER, L. 2018. Deep contextualized word representations. Proceedings of NAACL 2018.

RAJPURKAR, P. ZHANG, J.; LOPYREV, K., LIANG, P. **SQuAD: 100,000+ Questions for Machine Comprehension of Text**. . In: PROCEEDINGS OF THE 2016 CONFERENCE ON EMPIRICAL METHODS IN NATURAL LANGUAGE PROCESSING. nov. 2016. https://doi.org/10.18653/v1/D16-1264. Disponível em: <https://aclweb.org/anthology/papers/D/D16/D16-1264/>. Acesso em: 9 abr. 2019

REHUREK, R., SOJKA, P. Gensim–python framework for vector space modelling. NLP Centre, Faculty of Informatics, Masaryk University, Brno, Czech Republic, 2011.

RODRIGUES, J., BRANCO, A., NEALE, S., SILVA, J. LX-DSemVectors: Distributional Semantics Models for Portuguese. Computational Processing of the Portuguese Language: 12th International Conference (PROPOR-2016). Springer International Publishing, 2016.

RUDER, S. **Neural Transfer Learning for Natural Language Processing**. [s.l.] {National University of Ireland, Galway, 2019.

RUDER, S. **NLP's ImageNet moment has arrived.** 2018**.** Disponível em: <http://ruder.io/nlp-imagenet/>. Acesso em: 9 abr. 2019.

RUDER, S., VULIC, I., SØGAARD, A. A Survey of Cross-Lingual Word Embedding Models. The Journal of Artificial Intelligence Research, 2018. https://doi.org/10.17863/CAM.30462

SAHLGREN, M. 2008 The Distributional Hypothesis. Rivista di Linguistica (Italian Journal of Linguistics), 20. pp. 33-53.

SCHNABEL, T., LABUTOV, I., MIMNO, D., JOACHIMS, T. Evaluation methods for unsupervised word embeddings. EMNLP. 2015.

SHOHAM, Y., PERRAULT , R., BRYNJOLFSSON, E., CLARK, J., MANYIKA, J., NIEBLES, J., LYONS, T., ETCHEMENDY , J., GROSZ, B., BAUER, Z, "The AI Index 2018 Annual Report", AI Index Steering Committee, Human-Centered AI Initiative, Stanford University, Stanford, CA, December 2018.

SOCHER, R.; PERELYGIN, A.; WU, J., CHUANG, J., MANNING, C., NG, A., POTTS, C. Recursive Deep Models for Semantic Compositionality Over a Sentiment Treebank. In: Proceedings of the 2013 Conference on Empirical Methods in Natural Language Processing, 2013, p. 1631–1642.

SUTSKEVER, Ilya; VINYALS, Oriol; LE, Quoc V. Sequence to Sequence Learning with Neural Networks. In: **Proceedings of the 27th International Conference on Neural Information Processing Systems - Volume 2**. Cambridge, MA, USA: MIT Press, 2014, p. 3104–3112. (NIPS'14).

VAN DER MAATEN, L.J.P., HINTON, G. Visualizing High-Dimensional Data Using t-SNE. Journal of Machine Learning Research 9: pp. 2579-2605, 2008.

VASWANI, A. SHAZEER, N., PARMAR, N., USZKOREIT, J., JONES, L., GOMEZ, A., KAISER, L., POLOSUKHIN, I. Attention is All you Need. In: GUYON, I. et al. (Eds.). . Advances in Neural Information Processing Systems 30. Curran Associates, Inc., 2017. p. 5998–6008.

WU, Y., SCHUSTER, M., CHEN, Z., LE, Q.V., NOROUZI, M., MACHEREY, W., KRIKUN, M., CAO, Y., GAO, Q., KLINGNER, J., SHAH, A., JOHNSON, M., LIU, X., KAISER, L., GOUWS, S., KATO, Y., KUDO, T., KAZAWA, H., STEVENS, K., KURIAN, G., PATIL, N., WANG, W., YOUNG, C., SMITH, J., RIESA, J., RUDNICK, A., VINYALS, O., CORRADO, G.S., HUGHES, M., DEAN, J. (2016). Google's Neural Machine Translation System: Bridging the Gap between Human and Machine Translation. CoRR, abs/1609.08144.

YADAV, V., BETHARD, S. **A Survey on Recent Advances in Named Entity Recognition from Deep Learning models.** Proceedings of the 27th International Conference on Computational Linguistics, pages 2145–2158 Santa Fe, New Mexico, USA, August 20-26, 2018.

YIN, W.; KANN, K.; YU, M.; Schütze, H. Comparative Study of CNN and RNN for Natural Language Processing. arXiv:1702.01923, 2017. Disponível em: <http://arxiv.org/abs/1702.01923>. Acesso em: 15 abr. 2019.

YOUNG, T., HAZARIKA, D., PORIA, S., CAMBRIA, E. 2018. *Recent Trends in Deep Learning Based Natural Language Processing*. IEEE Computational Intelligence Magazine, vol.13. doi: 10.1109/MCI.2018.2840738

USP. Repositório de Word Embeddings do NILC. NILC - Núcleo Interinstitucional de Linguística Computacional. Disponível em: http://www.nilc.icmc.usp.br/embeddings

WORLD ECONOMIC FORUM. **Digital Transformation Initiative: Oil and Gas Industry.** 2017 [White paper]. Disponível em: < http://reports.weforum.org/digital-transformation/oil-gas//>. Acesso em: 15 abr. 2019.

ZHANG, L., WANG, S., LIU, B. (2018). **Deep Learning for Sentiment Analysis : A Survey**. Wiley Interdisciplinary Reviews: Data Mining and Knowledge Discovery. 10.1002/widm.1253. 201